\documentclass[letterpaper]{article} % DO NOT CHANGE THIS
\usepackage{aaai24}  % DO NOT CHANGE THIS
\usepackage{times}  % DO NOT CHANGE THIS
\usepackage{helvet}  % DO NOT CHANGE THIS
\usepackage{courier}  % DO NOT CHANGE THIS
\usepackage[hyphens]{url}  % DO NOT CHANGE THIS
\usepackage{graphicx} % DO NOT CHANGE THIS
\usepackage{natbib}  % DO NOT CHANGE THIS AND DO NOT ADD ANY OPTIONS TO IT
\usepackage{caption} % DO NOT CHANGE THIS AND DO NOT ADD ANY OPTIONS TO IT
\usepackage{algorithm}
\usepackage{algorithmic}
\usepackage{multirow}
\usepackage{booktabs}
\usepackage{subfigure}
\usepackage{amssymb}
\usepackage{bbding}
\usepackage{bm}
\usepackage{newfloat}
\usepackage{listings}
\DeclareCaptionStyle{ruled}{labelfont=normalfont,labelsep=colon,strut=off} % DO NOT CHANGE THIS
\floatstyle{ruled}
\newfloat{listing}{tb}{lst}{}
\floatname{listing}{Listing}
\title{CoIE: Chain-of-Instruct Editing for Multi-Attribute Face Manipulation
}
\author{
    Zhenduo Zhang,
    Bo-Wen Zhang,
    Guang Liu
}

\affiliations{
    Beijing Academy of Artificial Intelligence,  China \\
    \{zdzhang, bwzhang, liuguang\}@baai.ac.cn
}

\begin{document}

\maketitle

\begin{abstract}
Current text-to-image editing models often encounter challenges with smoothly manipulating multiple attributes using a single instruction. Taking inspiration from the Chain-of-Thought prompting technique utilized in language models, we present an innovative concept known as \textbf{Chain-of-Instruct Editing (CoIE)}, which enhances the capabilities of these models through step-by-step editing using a series of instructions. In particular, in the context of face manipulation, we leverage the contextual learning abilities of a pretrained Large Language Model (LLM), such as GPT-4, to generate a sequence of instructions from the original input, utilizing a purpose-designed 1-shot template. To further improve the precision of each editing step, we conduct fine-tuning on the editing models using our self-constructed instruction-guided face editing dataset, \textbf{Instruct-CelebA}. And additionally, we incorporate a super-resolution module to mitigate the adverse effects of editability and quality degradation. Experimental results across various challenging cases confirm the significant boost in multi-attribute facial image manipulation using chain-of-instruct editing. This is evident in enhanced editing success rates, measured by CLIPSim and Coverage metrics, improved by \textbf{17.86\%} and \textbf{85.45\%} respectively, and heightened controllability indicated by Preserve L1 and Quality metrics, improved by \textbf{11.58\%} and \textbf{4.93\%} respectively.

\end{abstract}

\begin{figure}[!htb]
    \includegraphics[width=1.0\linewidth]{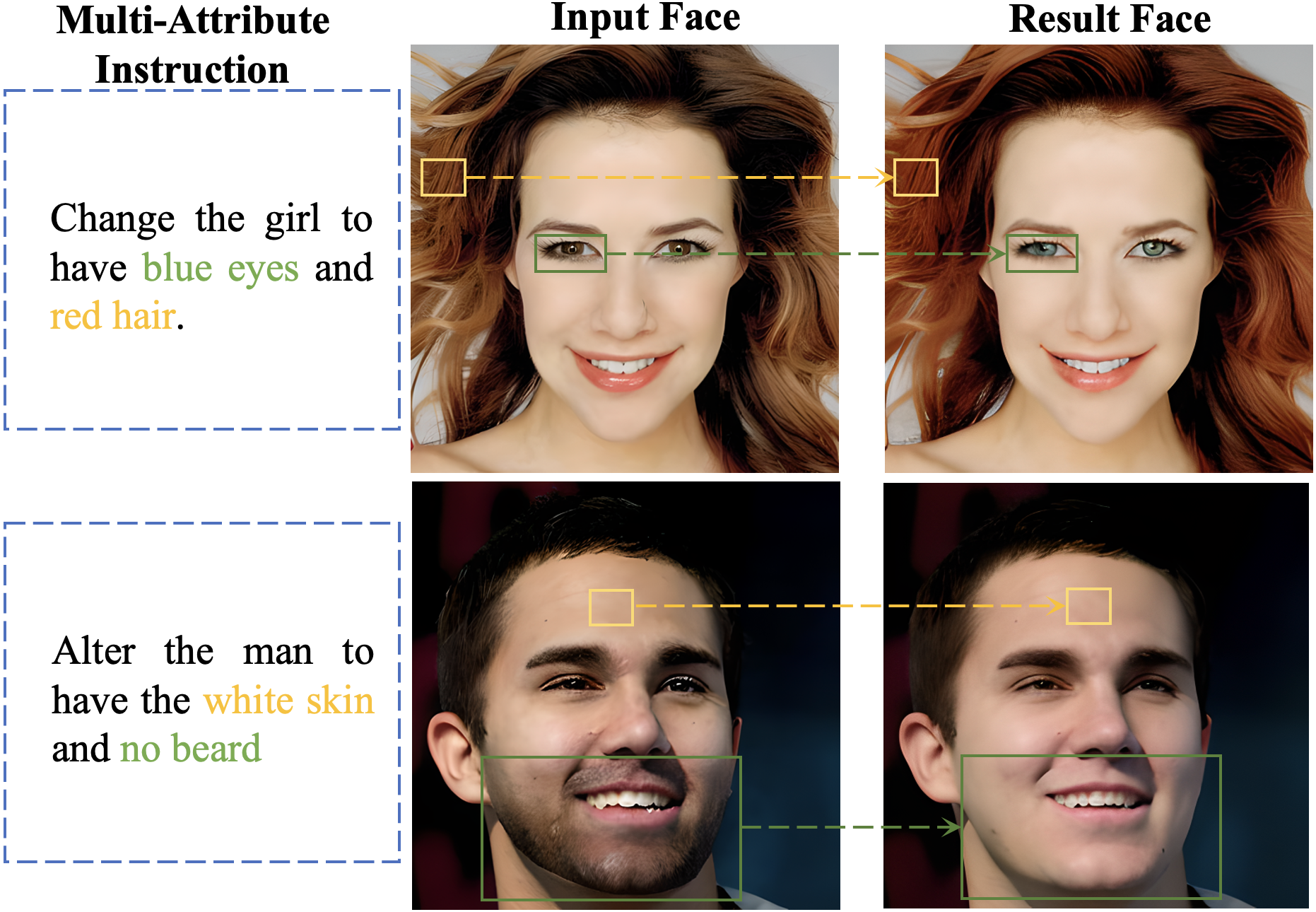}
    \captionof{figure}{Example of the Multiple Attribute Editing. Given one multi-attribute instruction and an input face image, the result face is generated according to the instruction description. The changes in the attributes are marked by rectangles in different colors.}
    \label{fig:mae}
\end{figure}%
\section{Introduction}
Text-guided Face editing holds notable potential for real-life applications, e.g., photography, advertising, and social media~\cite{chatface, ding2023diffusionrig}. Most works~\cite{10.1007/978-3-031-19784-0_41, p2p, Kawar2022ImagicTR, meng2022sdedit, instructpix2pix, zhang2023magicbrush} in text-guided image editing primarily focused on single-attribute editing scenarios, where each instruction modifies only one attribute of the image at a time. As we know, people often modify multiple face attributes to fulfill their necessities. Such modifications are usually guided by an instruction containing editing operations on multiple face attributes. However, current image editors struggle to handle instructions involving multiple attributes, leading to inconsistencies between the instructions and the result images. We term the task to edit multiple attributes using one instruction as \textbf{Multi-Attribute Editing}. Some examples of the Multi-Attribute Editing can be seen from Figure~\ref{fig:mae}, where each multi-attribute instruction contains two attribute changes and each result image are generated according to the instruction description.

In this paper, we introduce a simple yet effective approach \textbf{Chain-of-Instruct Editing (CoIE)} to alter the existing single attribute editor into a multiple attribute editor in a step-by-step manner.  We identify one main challenge in this alteration arises from the inability of current editors to precisely understand the instructions involving multiple attribute changes, leading to the neglect of necessary modifications. To address such a problem, we introduce a large language model(LLM)~\cite{openai2023gpt4} to decompose the compound instructions, involving multiple attribute modifications, into simpler single-attribute editing instructions. By doing so, multi-attribute editing can be accomplished by executing a sequence of single-attribute instructions instead of a single instruction involving multiple attributes. Therefore, we can utilize off-the-shelf single-attribute editors to accomplish multiple attribute editing in a step-by-step fashion. Besides, we observe that the consistency with the instructions and the quality of the generated images progressively deteriorate, as the number of consecutive edits increases. This editability and quality degradation is undesirable in the step-by-step editing processes in Chain-of-Instruct Editing. Therefore, we introduce an additional super-resolution module to recover detailed information about faces before editing them. It substantially enhances the editability and quality of the generated images during intermediate processes. In addition, we establish a large-scale dataset called \textbf{Instruct-CelebA} to enhance the controllability of Chain-of-Instruct Editing. By finetuning editing models on the Instruct-CelebA, we can significantly improve the consistency between the instruction and the result face and diminish irrelevant attribute modification. 

The evaluation way of the Multi-Attribute Editing problem has not been explored systematically up to now. In our work, we build a test dataset including 200 faces sampled from CelebAMask-HQ~\cite{lee2020maskgan} and each face has three kinds of instructions, which contain 2, 3, and 4 attribute changes. Besides, we systematically evaluate the performance of Multi-Attribute Editing from three dimensions: consistency with the instruction(CLIPSim~\cite{instructpix2pix}, Coverage), the preservation of the non-target region(Preserve L1), and the quality of the image(Quality Score~\cite{9156903}). Especially, the Coverage value counts the proportion of correctly modified attributes in the dataset to all attributes that need to be modified, and is an explicit metric to show the consistency between the result face and the instructions. The Preserve L1 is introduced by us to measure the controllability of the models from the granularity of the local region. 

To sum up, the primary contributions of this work are listed as follows:
\begin{itemize}

\item We propose Chain-of-Instruct Editing (CoIE) to accomplish multiple-face attribute editing by altering the existing single-attribute editor into a multiple-attribute editor.  This step-by-step approach improves the reasoning capacity of editing models when handling Multi-Attribute Editing problems. 

\item We introduce Instruct-CelebA, a large-scale training dataset for instruction-guided face editing. This dataset significantly enhances the controllability of face manipulation in Multi-Attribute Editing scenarios.

\item To mitigate the degradation of editability and quality during the step-by-step edits, we propose to incorporate an additional super-resolution module before the image editing model to further improve the performance in Multi-Attribute Editing scenarios.

\item We build a test dataset and a systematical evaluation way for the task of Multi-Attribute Editing by assessing its consistency with the instruction, preservation of the non-target region, and image quality. Through extensive experiments, we validate the effectiveness and superiority of our approach in comparison to existing methods.
\end{itemize}

\section{Related Work}

\subsection{Text-guided Image Editing}
% In general, image editing models perform editing tasks by either transferring styles~\cite{7780634} or translating between image domains~\cite{10.1007/978-3-030-01219-9_11, 8100115}.

Text-guided image editing aims to manipulate the image with the guidance of text description or text instructions. Based on CLIP~\cite{Radford2021LearningTV}, Many works focus on utilizing CLIP embeddings to guide the editing process~\cite{9879075, Crowson2022VQGANCLIPOD, 9879284, 9880189, 9710854, 10.1007/978-3-031-19784-0_41}. Recently, pretrained text-to-image diffusion models~\cite{ldm, Imagen} greatly facilitate the development of the image editing field~\cite{9879075, p2p, Kawar2022ImagicTR, meng2022sdedit, instructpix2pix}.Using cross-attention control, Prompt2Prompt~\cite{p2p} can perform local and global editing by modifying words in the original prompts. Null Text Inversion~\cite{nulltext} optimizes the inverted diffusion trajectory of the input image and can conduct real image editing with the Prompt2Prompt. Through optimizing a text embedding that aligns with the input image, Imagic~\cite{Kawar2022ImagicTR} generates different images for editing by interpolating them with the target description. ChatFace~\cite{chatface} develops an interactive system by combining the zero-shot reasoning ability of large language models to perform efficient manipulations in diffusion semantic latent space. InstructPix2Pix~\cite{instructpix2pix} leverages synthetic texts generated by finetuned GPT-3 and images by Prompt2Prompt and can edit real images in an instruction-guided manner. MagicBrush~\cite{zhang2023magicbrush} further establishes a large-scale, manually annotated dataset for instruction-guided real image editing and finetunes InstructPix2Pix on it.

\begin{figure*}
\centering
    \includegraphics[width=0.8\linewidth]{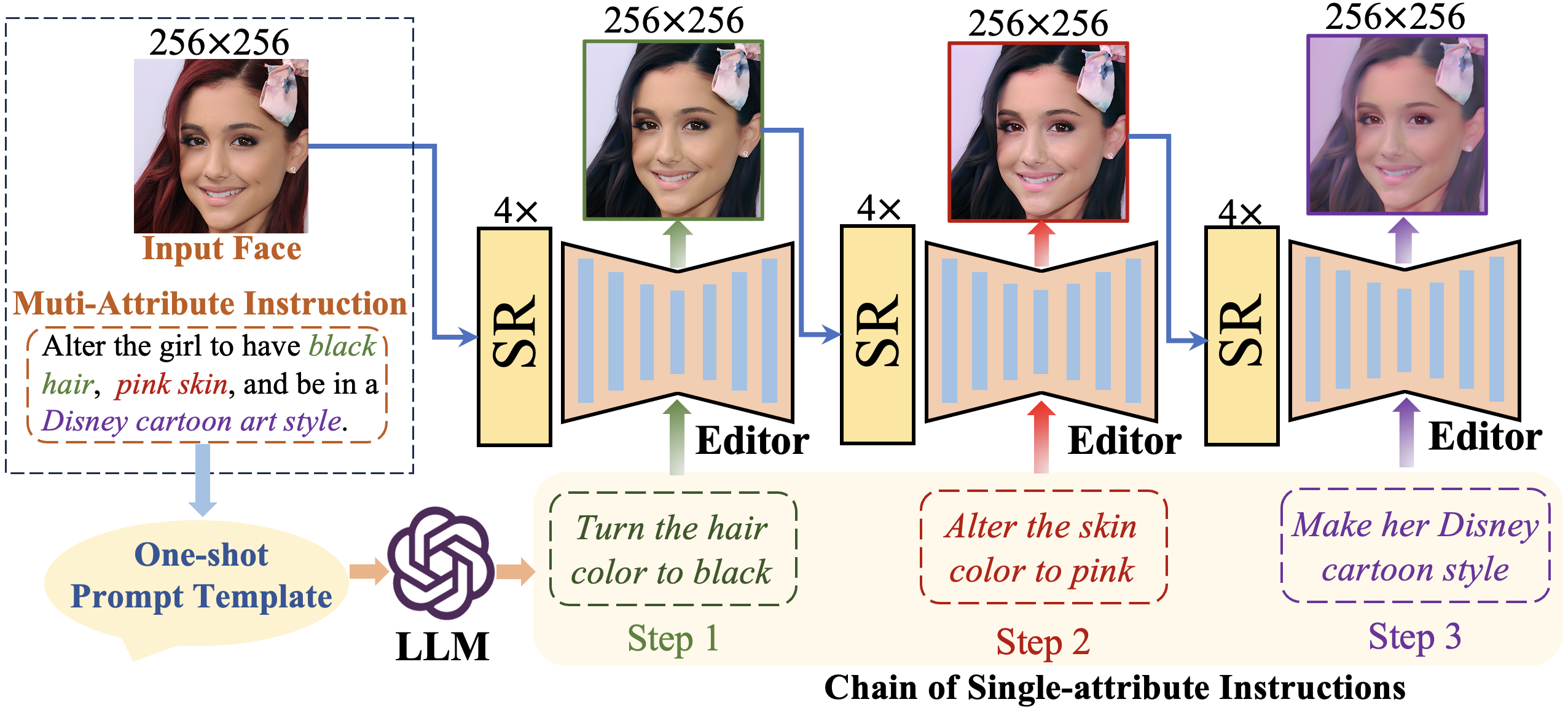}
    \captionof{figure}{
    Overall Framework. Given an instruction involving multiple attribute modifications, we adjoin it to a well-designed one-shot prompt template and feed the whole input to the LLM to perform instruction decomposition to generate a chain of single-attribute instructions. Then the input face will be edited in a step-by-step manner.}
    \label{fig:fig1}
\end{figure*}%

\subsection{Chain-of-Thought Reasoning of LLM}
The Chain-of-Thought (CoT)~\cite{wei2022chain} reasoning has been developed to boost the performance of large language models on challenging reasoning tasks, including arithmetic reasoning~\cite{DBLP:conf/acl/MiaoLS20}, commonsense reasoning~\cite{talmor-etal-2019-commonsenseqa}, and symbolic reasoning~\cite{wei2022chain}. The Chain-of-Thought (CoT) reasoning approach is commonly applied in two main settings when used with In-Context Learning - few-shot learning and zero-shot learning. Instead of feeding large language models standard question and answer examples, few-shot CoT learning works~\cite{wei2022chain, wang2023selfconsistency} provide LLMs with the step-by-step reasoning examples to facilitate models to generate a reasoning path decomposing the complex reasoning into multiple simpler steps. Unlike few-shot CoT, which includes human-provided task demonstrations in the prompts, zero-shot CoT does not have any annotated examples in the prompts. Zero-shot CoT directly generates the reasoning steps, then uses the produced Chain-of-Thought to deduce the answers. The zero-shot Chain-of-Thought approach was first introduced in the work~\cite{kojima2022large}, where the large language model is initially prompted with "Let's think step by step" to generate the reasoning steps. The model is then prompted with "Therefore, the answer is" to deduce the final answer.

\section{Overall Framework}
\subsection{Chain-of-Instruct Editing}
The Chain-of-Instruct Editing approach to address the challenge of Multi-Attribute Editing involving multiple attribute changes is depicted in Figure~\ref{fig:fig1}. Given an instruction $I$ containing $N$ attribute changes, we decompose the compound instruction into a chain of simpler single-attribute instructions, denoted as $ \left[I_1, I_2, ..., I_N \right]$. We denote the input image as $x_0$, then $x_n = \Phi(x_{n-1}, I_n)$ for $n=1,...,N$.

\begin{figure}
    \includegraphics[width=1.0\linewidth]{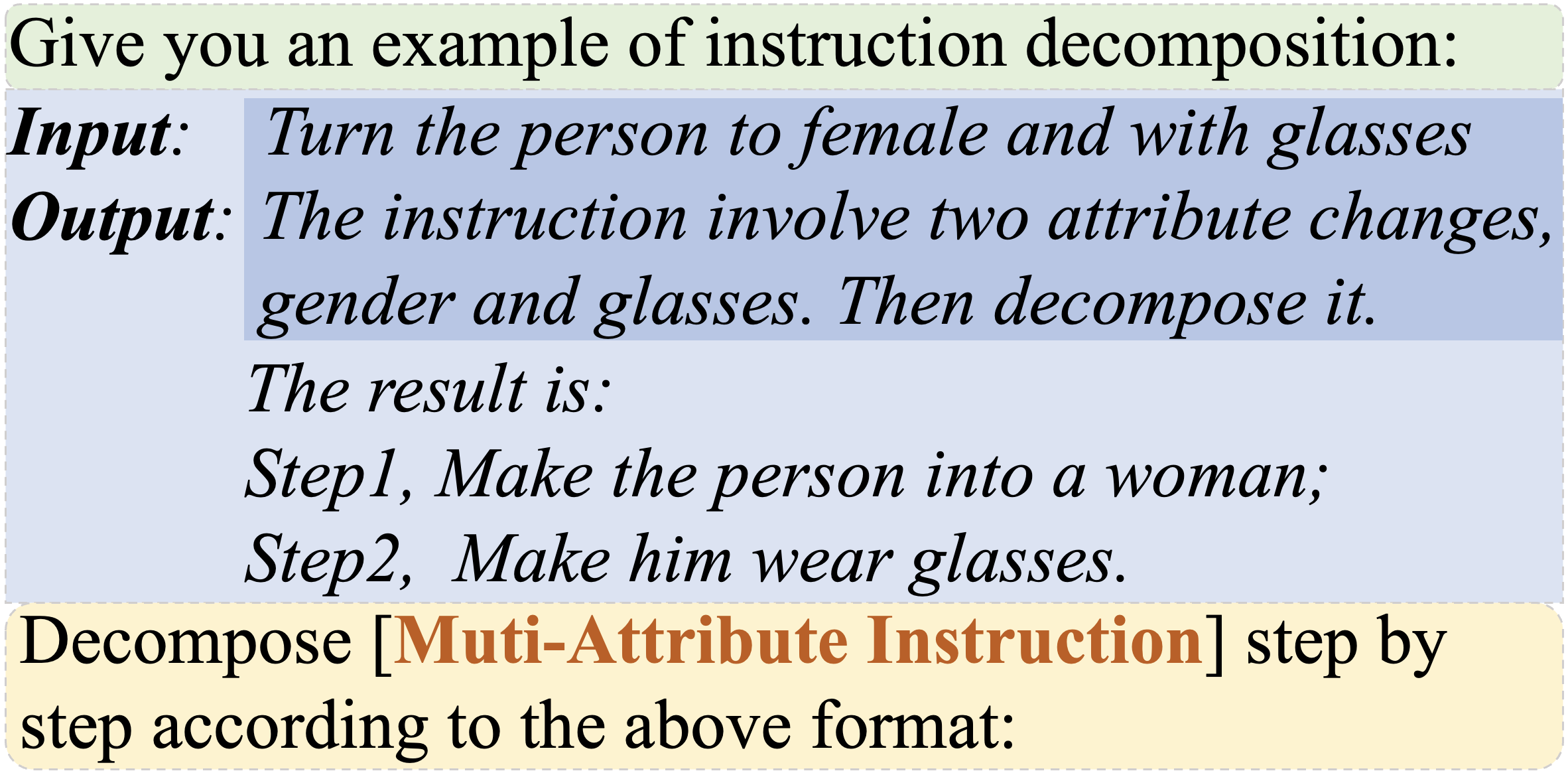}
    \captionof{figure}{Purpose-designed 1-shot prompt template.The one-shot prompt template consists of three components: Query (\textit{Yellow box}), Task Description (\textit{Green box}), and Demonstration (\textit{Blue box}).
    }
    \label{fig:prompt}
\end{figure}

One core procedure is to conduct the multi-attribute instruction decomposition, and we introduce a powerful few-shot learner, the LLM~\cite{openai2023gpt4}, to decompose the compound instruction $I$ into a chain of simpler single-attribute instructions $ \left[I_1, I_2, ..., I_N \right]$. The LLMs, such as GPT-4~\cite{openai2023gpt4}, have shown excellent performance in contextual learning, which can generalize to unseen tasks when providing a few of exemplars. In our approach, we design a purpose-designed one-shot prompt template to hint the GPT-4 to conduct instruction decomposition, and the one-shot prompt template for instruction decomposition is illustrated in Figure~\ref{fig:prompt}. The one-shot prompt template consists of three components: Query (\textit{Yellow box}), Task Description (\textit{Green box}), and Demonstration (\textit{Blue box}).  

The Task Description informs the large language models about the required task, with "Give you an example of instruction decomposition" indicating the task is instruction decomposition and hint the LLM to learn the correct decomposition manner from the examples we provide. The Demonstration provides an exemplar and sets constraints on the output format. A pair of input instruction and corresponding outputs is specified with the key words "Input" and "Output". Especially, we provide a hint of attribute recognition in the dark blue part: "The instruction involve two attribute changes, gender and glasses" and this hint helps the LLM to know which attributes are needed to decompose when processing a new multi-attribute instruction. In addition, the Demonstration constraints the output format and we only need to do little post-processing work to get the chain of the single-attribute instructions. The Query indicates the final task for the LLM, which is to decompose a multi-attribute instruction step by step according to the format given by the Demonstration. The input multi-attribute Instruction is incorporated in the prompt template. Denote the one-shot prompt template as $T$, then the decomposition process can be written as:$\left[I_1, I_2, ..., I_N \right] = LLM(I \oplus T)$.

\subsubsection{Image Editor}
It is essential to guarantee the controllability of the image editing model to improve the final performance of step-by-step editing. We consider the deficiency of controllability comes from the fact that the current training datasets, such as the dataset mentioned in InstructPix2Pix~\cite{instructpix2pix} and MagicBruth~\cite{zhang2023magicbrush}, lack massive high-quality face pairs before and after editing. In this paper, we try to improve the editing controllability of the model from the data perspective in our work. Based on CelebAMask-HQ~\cite{lee2020maskgan}, we build a large-scale dataset, Instruct-CelebA, which contains a number of high-quality triplets: $\left< instruction, input \ face, output\ face\right>$. In the Instruct-CelebA, the input face $F_{in}$ and the output face $F_{out}$ only differ in the target region which is specified in the instruction, and keep the non-target region unchanged. Such high-quality triplets guarantee the consistency between the instruction and the output face and improve the preservation of non-target regions, which should be kept in the editing operation. Trained on our Instruct-CelebA, the controllability of the editing model can be significantly improved, where the output face is modified within the target region and the change are consistent with the instruction.

\begin{figure}
  \centering
  \includegraphics[width=1.0\linewidth]{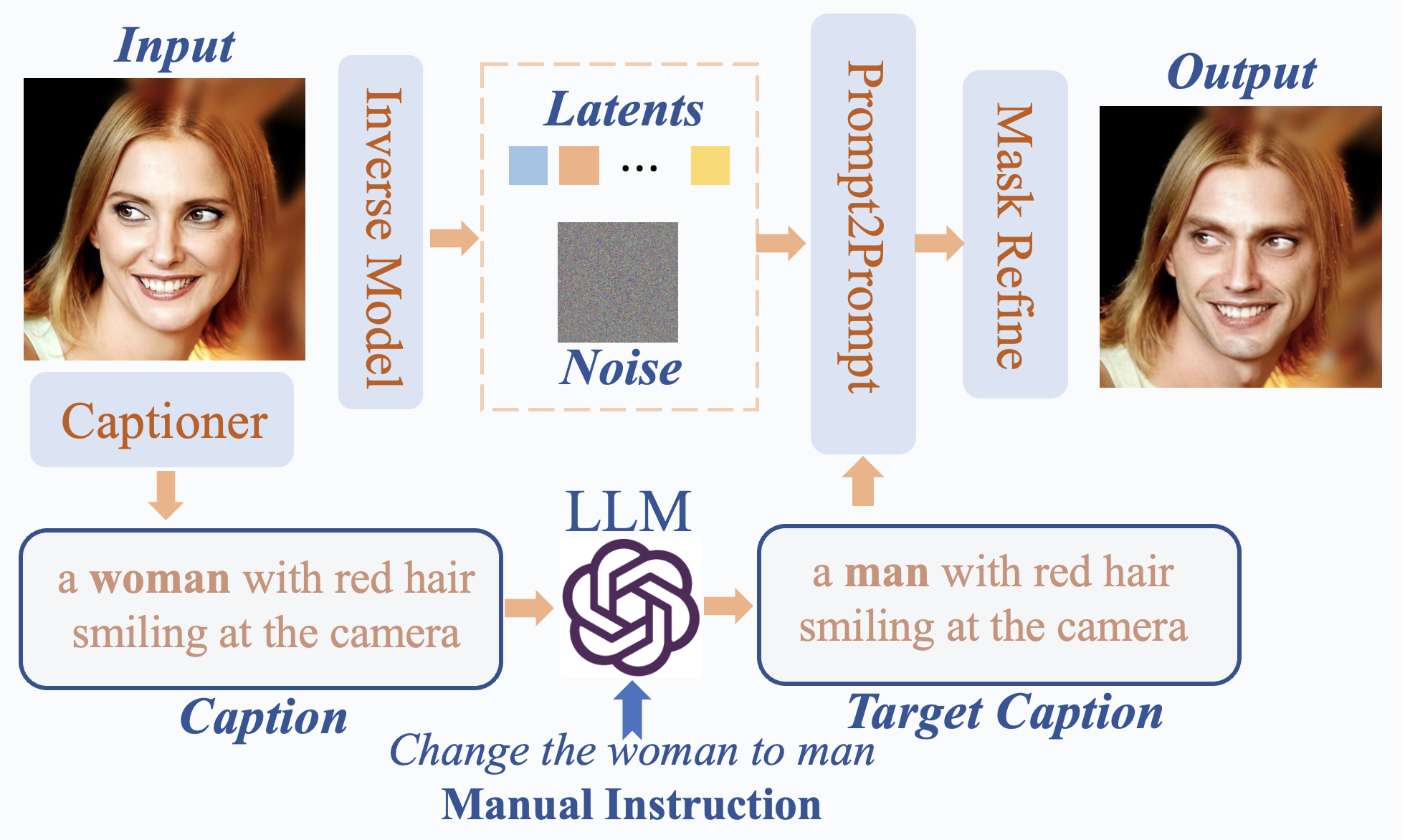}
  \caption{Data Annotation Pipeline}
  \label{fig:data}
\end{figure}

Next, we will briefly introduce the generation pipeline of face pairs, as is depicted in Figure~\ref{fig:data}. Given an input face image, we utilize one image caption model, such as BLIP2~\cite{li2023blip2}, to get the caption description. Then given the caption, we adopt the Null Text Inversion~\cite{nulltext} to invert the face to the noise and latents, which can reconstruct the input image using Stable Diffusion. We manually define some instruction templates, including nine edit types: hair, eyes, skin, gender, age, beard, glasses, anime style and expression. Similar to the way in InstructPix2Pix, we feed the manual instruction and the source caption to ChatGPT to generate the caption of the target face. Then, the Prompt2Prompt editing process can output the target face image with the input of noise, latents, source caption, and target caption. To further guarantee the preservation of the non-target region of the input face, we utilize the mask of the target region of editing to refine the output face. We denote the input face as $F_{in}$, the output image of Prompt2Prompt as $F_{p2p}$, and the target region mask as $I_{mask}$. The target region mask can be straightly selected from the mask annotations of CelebAMask-HQ~\cite{lee2020maskgan}. Then the final output face can be calculated as: $$F_{out} = F_{in} \times (1-I_{mask}) + F_{p2p} \times I_{mask}$$ To guarantee consistency with the target caption and the quality of faces, a CLIP~\cite{Radford2021LearningTV} model and a face quality assessment model~\cite{9156903} are deployed to filter the generated faces. 

As for the editing model selection, we reimplement the InstructPix2Pix by initializing it with the pre-trained AltDiffusion-M18\footnote{https://huggingface.co/BAAI/AltDiffusion-m18} weights and training it on the synthetically generated dataset released in the paper~\cite{instructpix2pix}. Then we further finetune it on our proposed Instruct-CelebA dataset to improve the controllability of face editing.

\subsubsection{Super-Resolution Model}
We observe that both the consistency with the instructions and the quality of the generated images gradually deteriorate during the consecutive editing process, negatively impacting the continuous editing process. Some works~\cite{Xu_2021_ICCV, 10.1145/3503161.3548134, chen2023dreamidentity} point that the degradation of editability may come from the loss of the detail information of generated images in the low level. This motivates us to recover detailed information about faces before editing them. Super-resolution models have been widely used to restore low-resolution images and are suitable to address our issue of detail information loss. Therefore, we propose a simple but effective way to address this issue: incorporating an additional super-resolution model before the editing model. In this work, RealESRGAN~\cite{wang2021realesrgan} is selected to conduct super-resolution, and it is denoted as $SR$. After introducing the super-resolution model, the Chain-of-Instruct Editing is formulated as $x_n = \Phi(SR(x_{n-1}), I_n)$ for $n=1,...,N$.

\begin{table}[!htb]
\small
\centering
\caption{
Summary of our Instruct-CelebA dataset. We summarize the number of face ids, the number of samples per attribute, and the type of changes of each attribute. 
}

\begin{tabular}{clccc}
\toprule
         
\textbf{Attribute} & \textbf{IDs}  &\textbf{Samples}  &\textbf{Attribute Changes}\\ \midrule
  
hair &10,345	& 40,139 & red/black/gray/blonde/brown \\             
skin &2,535	&   15,204  &  white/pink/brown/olive/...\\ 

eyes &3,838	& 7,676 & blue/black\\             
age &18,947	&  37,894 & younger/older\\ 

gender &10,266	&  20,532 & male/female\\             
anime &9,899	& 19,798 & real/anime\\

beard &2,002	& 4,004 & add/remove\\ 

glasses &4,294	&  8,588 & add/remove\\             
expression &7,624	& 27,947 & happy/angry/sad/fear/disgust
                                             
                         \\ \bottomrule
\end{tabular}
\label{tab:data_summary}
\vspace{-1em}
\end{table}

\begin{figure*}
  \centering
  \includegraphics[width=0.85\linewidth]{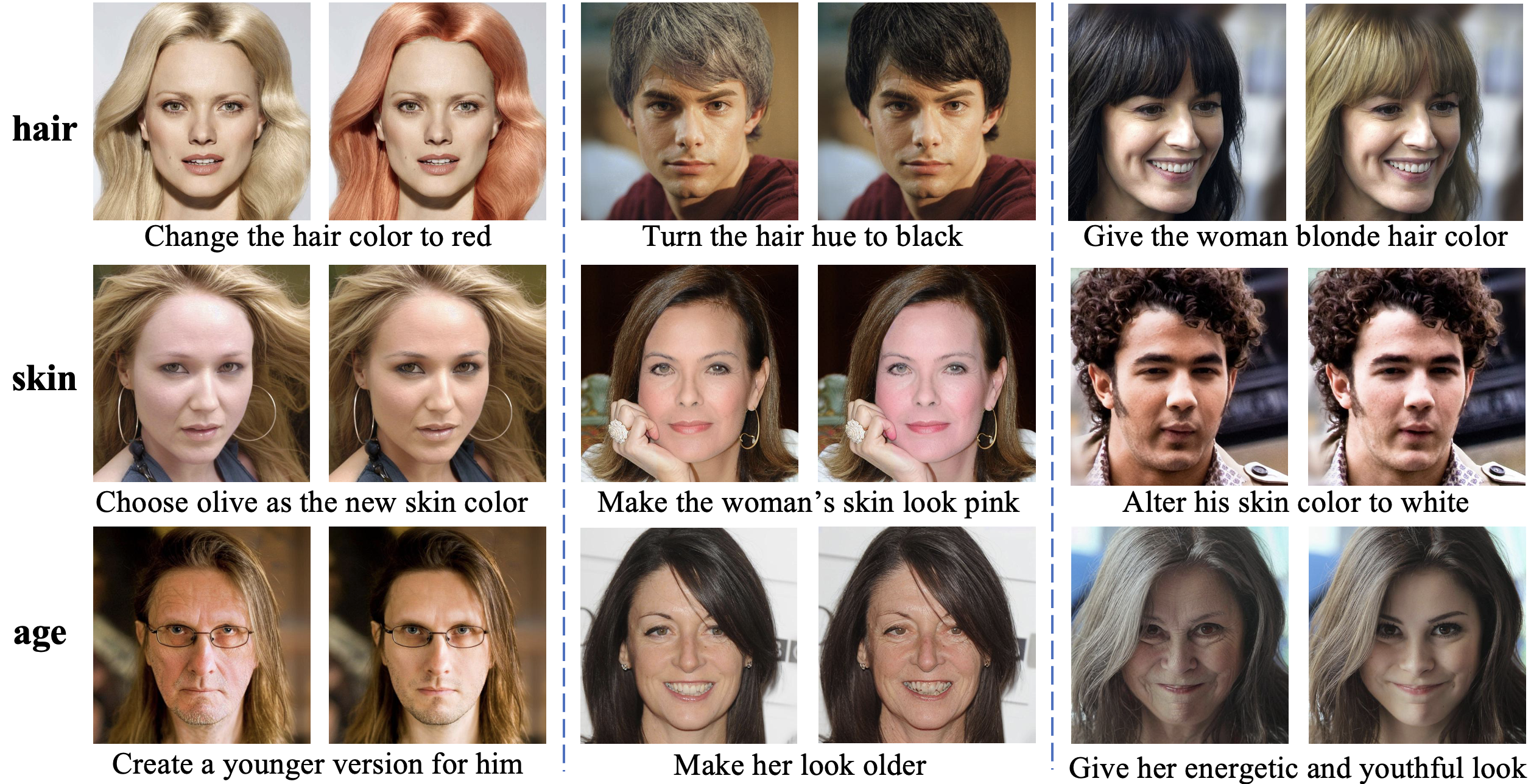}
  \caption{Examples of the Instruct-CelebA. The dataset contains massive triplets of input faces, output faces and instructions and each instruction only involves one attribute change.}
  \label{fig:data_example}
\end{figure*}

% \begin{figure}
%   \centering
%   \includegraphics[width=1.0\linewidth]{figure.png}
%   \caption{Editing performance comparison with the state-of-the-art methods on the different settings of attribute modification number in the instruction.}
%   \label{fig:fig_comparison}
% \end{figure}

\begin{table*}[t]
\small
\begin{center}
\caption{Performance of the Chain-of-Instruct Editing approach. We apply the CoIE approach to three different baselines: InstructPix2Pix, MagicBrush, and InstructPix2Pix*. We note that the InstructPix2Pix* is the reimplemented version of InstructPix2Pix using the AltDiffusion-M18 weights for the parameter initialization, finetuned on our Instruct-CelebA dataset, and inserted with a super-resolution module before the UNet. The CLIPSim and Coverage metrics are reported to show the improvement on the consistency with instructions. \textit{Note: The improvements of the metrics brought by COIE are listed in `()'.}}
\begin{tabular}{l|c|c|c|c|c|c}
\toprule
\multirow{3}{*}{\textbf{Models}} 
& \multicolumn{2}{c|}{\textbf{2 Attributes}} & \multicolumn{2}{c|}{\textbf{3 Attributes}} & \multicolumn{2}{c}{\textbf{4 Attributes}}  \\
\cmidrule{2-7}
& \textbf{CLIPSim$\uparrow$} & \textbf{Coverage$\uparrow$} & \textbf{CLIPSim$\uparrow$} & \textbf{Coverage$\uparrow$} & \textbf{CLIPSim$\uparrow$} & \textbf{Coverage$\uparrow$}  \\
\midrule
InstructPix2Pix~\cite{p2p} & 0.2291 & 0.535 & 0.2178 &  0.387 & 0.2164 & 0.368  \\
MagicBrush~\cite{zhang2023magicbrush} & 0.2276 & 0.480 & 0.2154 & 0.345 & 0.2160 & 0.355  \\
InstructPix2Pix* & 0.2406 & 0.563 & 0.2283 & 0.457 & 0.2258 & 0.383  \\
\midrule
\multirow{2}{*}{InstructPix2Pix + \textbf{CoIE}} & 0.2535 & 0.770 & 0.2352 &  0.665 & 0.2285 & 0.567  \\
                & \scriptsize{(+10.65\%)} & \scriptsize{(+43.92\%)} & \scriptsize{(+7.99\%)} & \scriptsize{(+71.83\%)} & \scriptsize{(+5.59\%)} & \scriptsize{(+54.08\%)}  \\
\multirow{2}{*}{MagicBrush + \textbf{CoIE}} & 0.2484 & 0.736 & 0.2346 & 0.652 & 0.2264 & 0.558  \\
                & \scriptsize{(+9.14\%)} & \scriptsize{(+53.33\%)} & \scriptsize{(+8.91\%)} & \scriptsize{(+88.98\%)} & \scriptsize{(+4.81\%)} & \scriptsize{(+57.18\%)}  \\
\multirow{2}{*}{InstructPix2Pix* + \textbf{CoIE} (Ours)} & \textbf{0.2642} & \textbf{0.835} & \textbf{0.2616} & \textbf{0.793} & \textbf{0.2557} & \textbf{0.719}  \\ 
& \scriptsize{(+9.81\%)} & \scriptsize{(+48.31\%)} & \scriptsize{(+14.59\%)} & \scriptsize{(+73.52\%)} & \scriptsize{(+13.24\%)} & \scriptsize{(+87.73\%)}  \\%
\bottomrule
\end{tabular}

\label{tb:CoIE}
\end{center}
\end{table*}

\begin{table*}[!htb]
\small
\centering
\caption{
Single-attribute editing result comparisons where each instruction only involves one attribute change. We evaluate our method across seven different attribute edits.
}

\begin{tabular}{cccccccccc}
\toprule
\textbf{Metrics}              & \textbf{w/o Instruct-CelebA}   & \textbf{hair}  &\textbf{eyes}	&\textbf{skin}	&\textbf{gender}	&\textbf{age}	&\textbf{anime}	&\textbf{expression} \\ \midrule
\multirow{2}{*}{\textbf{CLIPSim}$\uparrow$}    
                                              
                                     & $\bm{\times}$ &0.2684	&0.2454	&0.2539	&0.2286	&0.2465	&0.2315	&0.2306 \\
                                    & $\bm{\checkmark}$ &\textbf{0.2876}	&\textbf{0.2527}	&\textbf{0.2768}	&\textbf{0.2668}	&\textbf{0.2489}	&\textbf{0.2324}	&\textbf{0.2542}
                                                                                  \\ \midrule
\multirow{2}{*}{\textbf{Preserve L1}$\downarrow$}    
                                     & $\bm{\times}$ &16.958	&12.342	&17.136	&17.417	&13.209	&17.519	&11.118 \\
                                    & $\bm{\checkmark}$ &\textbf{11.628}	&\textbf{10.532}	&\textbf{8.568}	&\textbf{10.812}	&\textbf{9.231}	&\textbf{16.346}	&\textbf{10.251}

                         \\ \bottomrule
\end{tabular}
\label{tab:single-step}
\vspace{-1em}
\end{table*}

\section{Experiments}
\subsection{Dataset Analysis and Visualization}
We establish a large-scale dataset for instruction-guided face editing using the data annotation pipeline mentioned above. The statistic information is depicted in Table~\ref{tab:data_summary}. Our dataset supports nine facial editing attributes: hair, skin, eyes, age, gender, anime style, beard, glasses, and expression. The number of face ids and samples in each editing type are shown in the table, and we obtained 181,782 samples in total. We can sample the triplet of source-face, target-face, and editing instruction from the dataset to train the face editing model. Some examples are illustrated in Figure~\ref{fig:data_example}.

\subsection{Experiment Setup}

\subsubsection{Implement Details}
We select the InstructPix2Pix~\cite{p2p} architecture to build the image editing model. For future extension to multilingual usage, we implement a multilingual InstructPix2Pix via initializing it from the AltDiffusion-M18. pretrained weights and pretraining it on the released dataset in the original InstructPix2Pix paper. The pretraining stage contains 110000 training steps. After that, we futher finetune it on our Instruct-CelebA dataset for more 50000 steps. The optimizer is AdamW~\cite{loshchilov2019decoupled}. The learning rate is $1e-4$, with 5000 warmup steps on 32 NVIDIA A100-SXM4-40G GPUs. During the pretraining and finetuning stage, the parameters in text encoder is frozen and only the parameters in UNet is tuned. The batch size is 3072 and the input resolution of image is $256\times256$. To save the memory and speed up the training, the Xformer is adopted.
 
\subsubsection{Evaluation Metrics}
We systematically evaluate the performance of the Multi-Attribute Editing task from three aspects: consistency with the instruction, the preservation of the non-target region, and the quality of the image. The consistency with the instruction shows whether the attributes involved in the multi-attribute editing instruction are modified correctly, and it is the core metric to evaluate the performance improvement of the Chain-of-Instruct Editing approach. Given a multi-attribute instruction and the original caption of the input face, the caption of the result face can be generated via LLMs. Then the \textit{CLIP Similarity Score} between the result image and the result face caption is adopted to measure the consistency with the instruction. Besides, we further evaluate the consistency metric in a more straightforward manner via the \textit{Coverage} metric. The Coverage is defined as the proportion of correctly modified attributes in the dataset to all attributes that need to be modified. The larger CLIP Similarity Score and Coverage indicate better consistency performance. 

The preservation of the non-target region and the image quality measure the controllability of the models. Good preservation refers to that only the target region to be edited should be changed, and the non-target region should be the same as the source image. \textit{Preserve L1} is used to measure the preservation and is formulated as:
$$Preserve L1 = E[(I_{out} - I_{in})\times(1-I_{mask})]$$
The image quality is to measure the quality of the result face, and we adopt the SER-FIQ~\cite{9156903} to evaluate the face quality.

\subsubsection{Test Dataset Construction}
We randomly select 200 face samples with quality scores over $0.7$ from the original CelebAMask-HQ~\cite{lee2020maskgan} to form the test dataset. The reason we use samples from CelebAMask-HQ to evaluate our approach is that the dataset contains accurate masks of each attribute, such as hair, eyes, skin and etc. Using these attribute masks, we can easily evaluate the preservation of non-target regions to measure the controllability better. We establish a set of compound instructions involving $2$, $3$, and $4$ attribute changes for each sample in the test dataset.

\subsubsection{Performance Analysis of CoIE on Various Baselines}
The Chain-of-Instruct Editing alters the existing single-attribute editor into a multi-attribute editor in a step-by-step manner and facilitates the single-attribute editor to understand and execute the multi-attribute instructions better. In this part, we verify the assumption via adopting the CoIE to three baseline models, InstructPix2Pix~\cite{p2p}, MagicBrush~\cite{zhang2023magicbrush}, and InstructPix2Pix*, to show the improvement of the consistency between the result face and instructions. 

According to Table~\ref{tb:CoIE}, the CoIE approach can significantly improve both the CLIPSim and Coverage metrics on different settings of attribute numbers after employed to three baselines. For example, the InstructPix2Pix* gains an average improvement of $12.55\%$ on CLIPSim and $69.85\%$ on Coverage after adopting CoIE. We find there is a positive correlation between performance on Coverage and CLIPSim, and this proves that Coverage and CLIPSim can simultaneously illustrate the performance on editing consistency. Therefore, the improvements on three baselines verify the positive effect of the CoIE on enhancing the consistency between the result face and instructions. 

We observe the improvement in Coverage is much greater than that in CLIPSim. We attribute this issue to the fact that image captions contain rich entities such as "hair", "eyes", "women", etc. These entities give CLIPSim a higher starting value, resulting in a relatively small improvement in more accurate entity descriptions. Unlike the CLIPSim, Coverage directly reflects the performance of instruction execution by counting the proportion of correctly edited attributes among all attributes that need to be edited.

\textit{Notably, the current state-of-the-art model on the Multi-Attribute Face Editing task is the InstructPix2Pix without CoIE approach. As is shown in Table~\ref{tb:CoIE}, our model InstructPix2Pix* + CoIE significantly exceeds the existing model by a large margin and achieves the new state-of-the-art performance.}

\begin{table}[!htb]
\small
\centering
\caption{
Ablation study of the effect of our proposed Instruct-CelebA dataset. Four metrics(CLIPSim, Coverage, Preserve L1, and Quality) are utilized to measure the editing performance.
}

\begin{tabular}{clccc}
\toprule
\multirow{2}{*}{\textbf{Attributes}} & \multirow{2}{*}{\textbf{Metrics}}  &\multicolumn{2}{c}{\textbf{ w/o Instruct-CelebA}} \\
 \cline{3-4}
& & $\bm{\times}$ & $\bm{\checkmark}$ \\ \midrule
\multirow{4}{*}{\textbf{2}}    
                                     &CLIPSim &0.2568	&\textbf{0.2642}$\uparrow_{\mathbf{+2.88\%}}$  \\        &Coverage &0.806	&\textbf{0.835}$\uparrow_{\mathbf{+3.64\%}}$  \\    
                                     & Preserve L1 &11.781	&\textbf{10.608}$\downarrow_{\mathbf{-9.96\%}}$ \\
                                    & Quality &0.613    &\textbf{0.641}$\uparrow_{\mathbf{+4.57\%}}$
                                                                                  \\ \midrule
\multirow{4}{*}{\textbf{3}}    
                                       &CLIPSim &0.2523	&\textbf{0.2616}$\uparrow_{\mathbf{+3.69\%}}$  \\     
                                       &Coverage &0.757	&\textbf{0.793}$\uparrow_{\mathbf{+4.75\%}}$  \\
                                     & Preserve L1 &12.011	&\textbf{10.914}$\downarrow_{\mathbf{-9.13\%}}$ \\
                                    & Quality &0.597	&\textbf{0.629}$\uparrow_{\mathbf{+5.36\%}}$          
                                      \\ \midrule
\multirow{4}{*}{\textbf{4}}    
                                       &CLIPSim &0.2431	&\textbf{0.2557}$\uparrow_{\mathbf{+5.18\%}}$  \\        &Coverage &0.675	&\textbf{0.719}$\uparrow_{\mathbf{+6.48\%}}$  \\     
                                     & Preserve L1 &12.164	&\textbf{11.093}$\downarrow_{\mathbf{-8.80\%}}$ \\
                                    & Quality &0.588	&\textbf{0.619}$\uparrow_{\mathbf{+5.27\%}}$
                                             
                         \\ \bottomrule
\end{tabular}
\label{tab:Instruct-CelebA}
\vspace{-1em}
\end{table}

\begin{figure*}
  \centering
  \includegraphics[width=0.9\linewidth]{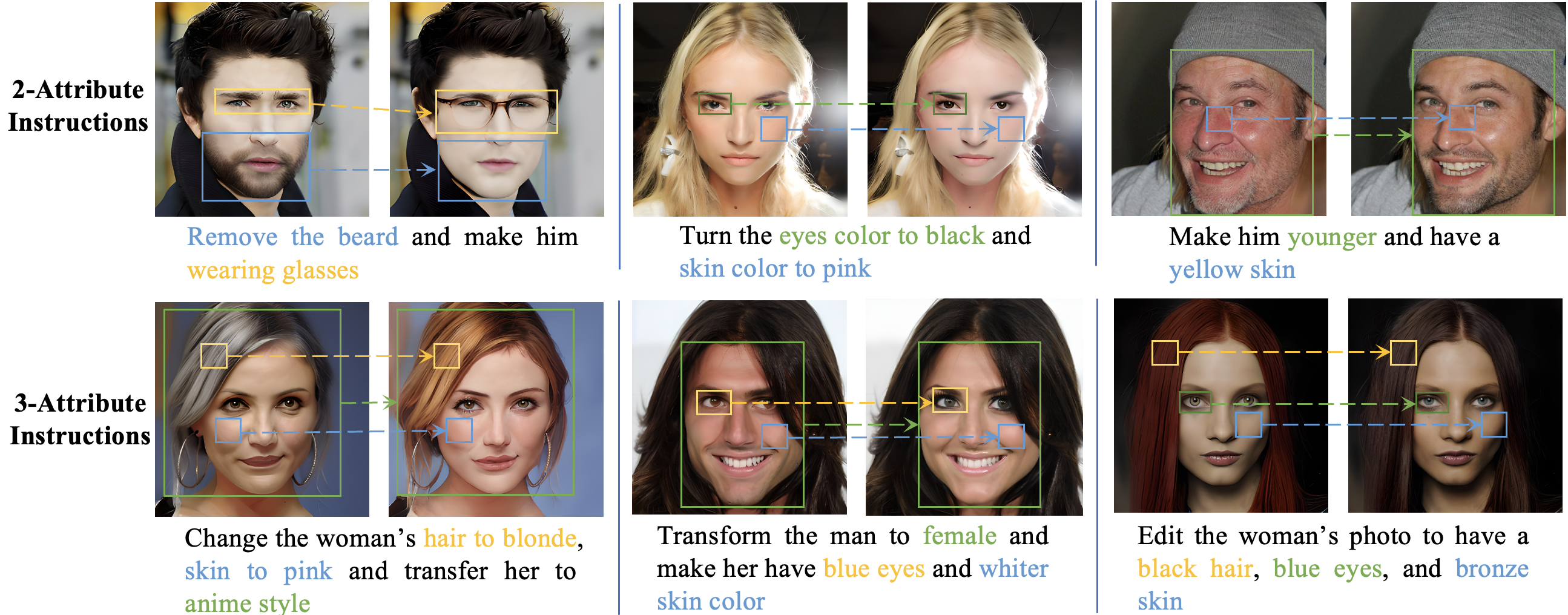}
  \caption{Visualization of our model's edit results. Given a multi-attribute instruction and a face to be edited, the modifications of different attributes are executed. In this figure, we show the cases of 2-attribute instructions and 3-attribute instructions.}
  \label{fig:visualization}
\end{figure*}

\begin{table}[!htb]
\small
\centering
\caption{
Ablation study of the effect of super-resolution model. Four metrics(CLIPSim, Coverage, Preserve L1, and Quality) are utilized to measure the editing performance.
}

\begin{tabular}{clccc}
\toprule
\multirow{2}{*}{\textbf{Attributes}} & \multirow{2}{*}{\textbf{Metrics}}  &\multicolumn{2}{c}{\textbf{w/o Super Resolution}} \\

\cline{3-4}
& & $\bm{\times}$ & $\bm{\checkmark}$ \\ \midrule

\multirow{3}{*}{\textbf{2}}    
                                     &CLIPSim &0.2460	&\textbf{0.2642}$\uparrow_{\mathbf{+7.40\%}}$  \\        &Coverage &0.782	&\textbf{0.835}$\uparrow_{\mathbf{+6.78\%}}$  \\  
                                     & Preserve L1 &11.652	&\textbf{10.608}$\downarrow_{\mathbf{-8.96\%}}$ \\
                                    & Quality &0.615    &\textbf{0.641}$\uparrow_{\mathbf{+4.23\%}}$
                                                                                  \\ \midrule
\multirow{3}{*}{\textbf{3}}    
                                       &CLIPSim &0.2363	&\textbf{0.2616}$\uparrow_{\mathbf{+10.71\%}}$  \\     &Coverage &0.730	&\textbf{0.793}$\uparrow_{\mathbf{+8.62\%}}$  \\    
                                     & Preserve L1 &12.347	&\textbf{10.914}$\downarrow_{\mathbf{-11.61\%}}$ \\
                                    & Quality &0.591	&\textbf{0.629}$\uparrow_{\mathbf{+6.43\%}}$          
                                      \\ \midrule
\multirow{3}{*}{\textbf{4}}    
                                       &CLIPSim &0.2267	&\textbf{0.2557}$\uparrow_{\mathbf{+12.79\%}}$  \\     &Coverage &0.642	&\textbf{0.719}$\uparrow_{\mathbf{+11.94\%}}$  \\      
                                     & Preserve L1 &13.492	&\textbf{11.093}$\downarrow_{\mathbf{-17.78\%}}$ \\
                                    & Quality &0.574	&\textbf{0.619}$\uparrow_{\mathbf{+7.84\%}}$
                                             
                         \\ \bottomrule
\end{tabular}
\label{tab:SR}
\vspace{-1em}
\end{table}

% ~\\
\subsubsection{Performance Analysis of the Instruct-CelebA Dataset}
The Instruct-CelebA dataset is established to guarantee the controllability of single-attribute editing and further improve the performance of Multi-Attribute Editing. We verify the effect of the Instruct-CelebA dataset in this part, and the ablation results are in Table~\ref{tab:single-step} and Table~\ref{tab:Instruct-CelebA}. Table~\ref{tab:single-step} shows the single-attribute editing comparison where each instruction only involves one attribute change, such as hair, eyes, etc. We observe that the CLIPSim and Preserve L1 can be improved across seven different attribute edits utilizing the Instruct-CelebA dataset, and our InstructFace also significantly exceeds the InstructPix2Pix in the single-attribute editing scenario. This indicates that our Instruct-CelebA dataset can help promote the consistency and controllability of single-attribute editing. Furthermore, we also verify the improvement in the setting of multiple attributes, as is depicted in Table~\ref{tab:Instruct-CelebA}. In different settings of attribute numbers, all the metrics(CLIPSim, Coverage, Preserve L1, and Quality) can be enhanced notably by a large margin compared with the ones without Instruct-CelebA. Both the results of the single-attribue setting and multi-attribute setting prove the effect and necessity of our Instruct-CelebA dataset.

\subsubsection{Performance Analysis of Super-Resolution Model}
As is illustrated in Table~\ref{tab:SR}, to verify the necessity of the Super-Resolution model, we conduct ablation experiments to compare the editing performance before and after adding the super-resolution model. We especially compute the percentage of improvement in CLIPSim, Coverage, Preserve L1, and Quality across different settings of attribute number in compound instructions. We have two key observations: 1. the super-resolution model can significantly enhance performance in all metrics across different settings of attribute numbers. 2. With the complexity of the instruction increasing, i.e., the more attributes, the improvement percentage of the four metrics becomes larger. For example, the improvement percentages of the CLIPSim are $7.40\%$, $10.71\%$, and $12.79\%$ correspondingly. The two observations prove that the super-resolution model can help eliminate the deterioration of consistency with the instructions, out-target part preservation, and the quality of the generated images during the consecutive editing process.

\subsubsection{Visualization Results}
We show some multi-attribute editing results in Figure~\ref{fig:visualization}. The various attributes are highlighted with different colors in the multi-attribute instructions. From this figure, we know that the attributes need to modified are correctly changed. The visualization results help prove the effectiveness of our CoIE approach.

\section{Conclusion}
We first propose a novel approach Chain-of-Instruct Editing(CoIE) to address the Multi-Attribute Editing problem. We alter the current single-attribute editor to solve the Multi-Attribute Editing task in a step-by-step way, by introducing the LLM to decompose the multi-attribute instruction into a chain of single-attribute instructions and executing them gradually. Secondly, we release a large-scale dataset called Instruct-CelebA for public usage for instruction-guided face editing, which can help improve the controllability of Multi-Attribute Editing. Thirdly, we propose to incorporate an additional super-resolution module before the image editing model to eliminate the degradation of editability and quality during the step-by-step edits. Finally, we systematically validate the effectiveness and superiority of our approach.

\bibliography{aaai24}

\begin{thebibliography}{31}
\providecommand{\natexlab}[1]{#1}

\bibitem[{Avrahami, Lischinski, and Fried(2022)}]{9879075}
Avrahami, O.; Lischinski, D.; and Fried, O. 2022.
\newblock Blended Diffusion for Text-driven Editing of Natural Images.
\newblock In \emph{2022 IEEE/CVF Conference on Computer Vision and Pattern
  Recognition (CVPR)}, 18187--18197.

\bibitem[{Bar-Tal et~al.(2022)Bar-Tal, Ofri-Amar, Fridman, Kasten, and
  Dekel}]{10.1007/978-3-031-19784-0_41}
Bar-Tal, O.; Ofri-Amar, D.; Fridman, R.; Kasten, Y.; and Dekel, T. 2022.
\newblock Text2LIVE: Text-Driven Layered Image and Video Editing.
\newblock In \emph{Computer Vision -- ECCV 2022}, 707--723. Cham: Springer
  Nature Switzerland.

\bibitem[{Brooks, Holynski, and Efros(2023)}]{instructpix2pix}
Brooks, T.; Holynski, A.; and Efros, A.~A. 2023.
\newblock InstructPix2Pix: Learning To Follow Image Editing Instructions.
\newblock In \emph{Proceedings of the IEEE/CVF Conference on Computer Vision
  and Pattern Recognition (CVPR)}, 18392--18402.

\bibitem[{Chen et~al.(2023)Chen, Fang, Liu, He, Huang, Zhang, and
  Mao}]{chen2023dreamidentity}
Chen, Z.; Fang, S.; Liu, W.; He, Q.; Huang, M.; Zhang, Y.; and Mao, Z. 2023.
\newblock DreamIdentity: Improved Editability for Efficient Face-identity
  Preserved Image Generation.
\newblock arXiv:2307.00300.

\bibitem[{Crowson et~al.(2022)Crowson, Biderman, Kornis, Stander, Hallahan,
  Castricato, and Raff}]{Crowson2022VQGANCLIPOD}
Crowson, K.; Biderman, S.~R.; Kornis, D.; Stander, D.; Hallahan, E.;
  Castricato, L.; and Raff, E. 2022.
\newblock VQGAN-CLIP: Open Domain Image Generation and Editing with Natural
  Language Guidance.
\newblock In \emph{European Conference on Computer Vision}.

\bibitem[{Ding et~al.(2023)Ding, Zhang, Xia, Jebe, Tu, and
  Zhang}]{ding2023diffusionrig}
Ding, Z.; Zhang, X.; Xia, Z.; Jebe, L.; Tu, Z.; and Zhang, X. 2023.
\newblock DiffusionRig: Learning Personalized Priors for Facial Appearance
  Editing.
\newblock arXiv:2304.06711.

\bibitem[{Hertz et~al.(2023)Hertz, Mokady, Tenenbaum, Aberman, Pritch, and
  Cohen-or}]{p2p}
Hertz, A.; Mokady, R.; Tenenbaum, J.; Aberman, K.; Pritch, Y.; and Cohen-or, D.
  2023.
\newblock Prompt-to-Prompt Image Editing with Cross-Attention Control.
\newblock In \emph{The Eleventh International Conference on Learning
  Representations}.

\bibitem[{Kawar et~al.(2023)Kawar, Zada, Lang, Tov, Chang, Dekel, Mosseri, and
  Irani}]{Kawar2022ImagicTR}
Kawar, B.; Zada, S.; Lang, O.; Tov, O.; Chang, H.-T.; Dekel, T.; Mosseri, I.;
  and Irani, M. 2023.
\newblock Imagic: Text-Based Real Image Editing with Diffusion Models.
\newblock In \emph{2023 IEEE/CVF Conference on Computer Vision and Pattern
  Recognition (CVPR)}.

\bibitem[{Kim, Kwon, and Ye(2022)}]{9879284}
Kim, G.; Kwon, T.; and Ye, J.~C. 2022.
\newblock DiffusionCLIP: Text-Guided Diffusion Models for Robust Image
  Manipulation.
\newblock In \emph{2022 IEEE/CVF Conference on Computer Vision and Pattern
  Recognition (CVPR)}, 2416--2425.

\bibitem[{Kojima et~al.(2022)Kojima, Gu, Reid, Matsuo, and
  Iwasawa}]{kojima2022large}
Kojima, T.; Gu, S.~S.; Reid, M.; Matsuo, Y.; and Iwasawa, Y. 2022.
\newblock Large Language Models are Zero-Shot Reasoners.
\newblock In Oh, A.~H.; Agarwal, A.; Belgrave, D.; and Cho, K., eds.,
  \emph{Advances in Neural Information Processing Systems}.

\bibitem[{Kwon and Ye(2022)}]{9880189}
Kwon, G.; and Ye, J.~C. 2022.
\newblock CLIPstyler: Image Style Transfer with a Single Text Condition.
\newblock In \emph{2022 IEEE/CVF Conference on Computer Vision and Pattern
  Recognition (CVPR)}, 18041--18050.

\bibitem[{Lee et~al.(2020)Lee, Liu, Wu, and Luo}]{lee2020maskgan}
Lee, C.-H.; Liu, Z.; Wu, L.; and Luo, P. 2020.
\newblock MaskGAN: Towards Diverse and Interactive Facial Image Manipulation.
\newblock arXiv:1907.11922.

\bibitem[{Li et~al.(2023)Li, Li, Savarese, and Hoi}]{li2023blip2}
Li, J.; Li, D.; Savarese, S.; and Hoi, S. 2023.
\newblock BLIP-2: Bootstrapping Language-Image Pre-training with Frozen Image
  Encoders and Large Language Models.
\newblock arXiv:2301.12597.

\bibitem[{Loshchilov and Hutter(2019)}]{loshchilov2019decoupled}
Loshchilov, I.; and Hutter, F. 2019.
\newblock Decoupled Weight Decay Regularization.
\newblock arXiv:1711.05101.

\bibitem[{Mao et~al.(2022)Mao, Cao, Gnanha, Yang, Li, and
  Ji}]{10.1145/3503161.3548134}
Mao, X.; Cao, L.; Gnanha, A.~T.; Yang, Z.; Li, Q.; and Ji, R. 2022.
\newblock Cycle Encoding of a StyleGAN Encoder for Improved Reconstruction and
  Editability.
\newblock In \emph{Proceedings of the 30th ACM International Conference on
  Multimedia}, MM '22, 2032–2041. New York, NY, USA: Association for
  Computing Machinery.
\newblock ISBN 9781450392037.

\bibitem[{Meng et~al.(2022)Meng, He, Song, Song, Wu, Zhu, and
  Ermon}]{meng2022sdedit}
Meng, C.; He, Y.; Song, Y.; Song, J.; Wu, J.; Zhu, J.-Y.; and Ermon, S. 2022.
\newblock {SDE}dit: Guided Image Synthesis and Editing with Stochastic
  Differential Equations.
\newblock In \emph{International Conference on Learning Representations}.

\bibitem[{Miao, Liang, and Su(2020)}]{DBLP:conf/acl/MiaoLS20}
Miao, S.; Liang, C.; and Su, K. 2020.
\newblock A Diverse Corpus for Evaluating and Developing English Math Word
  Problem Solvers.
\newblock In \emph{Proceedings of the 58th Annual Meeting of the Association
  for Computational Linguistics, {ACL} 2020, Online, July 5-10, 2020},
  975--984. Association for Computational Linguistics.

\bibitem[{Mokady et~al.(2023)Mokady, Hertz, Aberman, Pritch, and
  Cohen-Or}]{nulltext}
Mokady, R.; Hertz, A.; Aberman, K.; Pritch, Y.; and Cohen-Or, D. 2023.
\newblock Null-text Inversion for Editing Real Images using Guided Diffusion
  Models.
\newblock In \emph{2023 IEEE/CVF Conference on Computer Vision and Pattern
  Recognition (CVPR)}.

\bibitem[{OpenAI(2023)}]{openai2023gpt4}
OpenAI. 2023.
\newblock GPT-4 Technical Report.
\newblock arXiv:2303.08774.

\bibitem[{Patashnik et~al.(2021)Patashnik, Wu, Shechtman, Cohen-Or, and
  Lischinski}]{9710854}
Patashnik, O.; Wu, Z.; Shechtman, E.; Cohen-Or, D.; and Lischinski, D. 2021.
\newblock StyleCLIP: Text-Driven Manipulation of StyleGAN Imagery.
\newblock In \emph{2021 IEEE/CVF International Conference on Computer Vision
  (ICCV)}, 2065--2074.

\bibitem[{Radford et~al.(2021)Radford, Kim, Hallacy, Ramesh, Goh, Agarwal,
  Sastry, Askell, Mishkin, Clark, Krueger, and
  Sutskever}]{Radford2021LearningTV}
Radford, A.; Kim, J.~W.; Hallacy, C.; Ramesh, A.; Goh, G.; Agarwal, S.; Sastry,
  G.; Askell, A.; Mishkin, P.; Clark, J.; Krueger, G.; and Sutskever, I. 2021.
\newblock Learning Transferable Visual Models From Natural Language
  Supervision.
\newblock In \emph{International Conference on Machine Learning}.

\bibitem[{Rombach et~al.(2022)Rombach, Blattmann, Lorenz, Esser, and
  Ommer}]{ldm}
Rombach, R.; Blattmann, A.; Lorenz, D.; Esser, P.; and Ommer, B. 2022.
\newblock High-Resolution Image Synthesis with Latent Diffusion Models.
\newblock In \emph{2022 IEEE/CVF Conference on Computer Vision and Pattern
  Recognition (CVPR)}, 10674--10685.

\bibitem[{Saharia et~al.(2022)Saharia, Chan, Saxena, Li, Whang, Denton,
  Ghasemipour, Ayan, Mahdavi, Lopes, Salimans, Ho, Fleet, and Norouzi}]{Imagen}
Saharia, C.; Chan, W.; Saxena, S.; Li, L.; Whang, J.; Denton, E.~L.;
  Ghasemipour, S. K.~S.; Ayan, B.~K.; Mahdavi, S.~S.; Lopes, R.~G.; Salimans,
  T.; Ho, J.; Fleet, D.~J.; and Norouzi, M. 2022.
\newblock Photorealistic Text-to-Image Diffusion Models with Deep Language
  Understanding.
\newblock \emph{ArXiv}, abs/2205.11487.

\bibitem[{Talmor et~al.(2019)Talmor, Herzig, Lourie, and
  Berant}]{talmor-etal-2019-commonsenseqa}
Talmor, A.; Herzig, J.; Lourie, N.; and Berant, J. 2019.
\newblock {C}ommonsense{QA}: A Question Answering Challenge Targeting
  Commonsense Knowledge.
\newblock In \emph{Proceedings of the 2019 Conference of the North {A}merican
  Chapter of the Association for Computational Linguistics: Human Language
  Technologies, Volume 1 (Long and Short Papers)}, 4149--4158. Association for
  Computational Linguistics.

\bibitem[{Terhörst et~al.(2020)Terhörst, Kolf, Damer, Kirchbuchner, and
  Kuijper}]{9156903}
Terhörst, P.; Kolf, J.~N.; Damer, N.; Kirchbuchner, F.; and Kuijper, A. 2020.
\newblock SER-FIQ: Unsupervised Estimation of Face Image Quality Based on
  Stochastic Embedding Robustness.
\newblock In \emph{2020 IEEE/CVF Conference on Computer Vision and Pattern
  Recognition (CVPR)}, 5650--5659.

\bibitem[{Wang et~al.(2023)Wang, Wei, Schuurmans, Le, Chi, Narang, Chowdhery,
  and Zhou}]{wang2023selfconsistency}
Wang, X.; Wei, J.; Schuurmans, D.; Le, Q.~V.; Chi, E.~H.; Narang, S.;
  Chowdhery, A.; and Zhou, D. 2023.
\newblock Self-Consistency Improves Chain of Thought Reasoning in Language
  Models.
\newblock In \emph{The Eleventh International Conference on Learning
  Representations}.

\bibitem[{Wang et~al.(2021)Wang, Xie, Dong, and Shan}]{wang2021realesrgan}
Wang, X.; Xie, L.; Dong, C.; and Shan, Y. 2021.
\newblock Real-ESRGAN: Training Real-World Blind Super-Resolution with Pure
  Synthetic Data.
\newblock arXiv:2107.10833.

\bibitem[{Wei et~al.(2022)Wei, Wang, Schuurmans, Bosma, brian ichter, Xia, Chi,
  Le, and Zhou}]{wei2022chain}
Wei, J.; Wang, X.; Schuurmans, D.; Bosma, M.; brian ichter; Xia, F.; Chi,
  E.~H.; Le, Q.~V.; and Zhou, D. 2022.
\newblock Chain of Thought Prompting Elicits Reasoning in Large Language
  Models.
\newblock In \emph{Advances in Neural Information Processing Systems}.

\bibitem[{Xu et~al.(2021)Xu, Du, Xiao, Xu, and He}]{Xu_2021_ICCV}
Xu, Y.; Du, Y.; Xiao, W.; Xu, X.; and He, S. 2021.
\newblock From Continuity to Editability: Inverting GANs With Consecutive
  Images.
\newblock In \emph{Proceedings of the IEEE/CVF International Conference on
  Computer Vision (ICCV)}, 13910--13918.

\bibitem[{Yue et~al.(2023)Yue, Guo, Ning, Cui, Zhu, and Yuan}]{chatface}
Yue, D.; Guo, Q.; Ning, M.; Cui, J.; Zhu, Y.; and Yuan, L. 2023.
\newblock ChatFace: Chat-Guided Real Face Editing via Diffusion Latent Space
  Manipulation.
\newblock arXiv:2305.14742.

\bibitem[{Zhang et~al.(2023)Zhang, Mo, Chen, Sun, and Su}]{zhang2023magicbrush}
Zhang, K.; Mo, L.; Chen, W.; Sun, H.; and Su, Y. 2023.
\newblock MagicBrush: A Manually Annotated Dataset for Instruction-Guided Image
  Editing.
\newblock arXiv:2306.10012.

\end{thebibliography}

\end{document}